# StreetviewLLM: Extracting Geographic Information Using a Chain-of-Thought Multimodal Large Language Model

Zongrong Li[a], Junhao Xu[b,c], Siqin Wang[a*], Yifan Wu[d], Haiyang Li[a]

[a] Spatial Sciences Institute, University of Southern California, Los Angeles, United States

[b] University of Malaya, Kuala Lumpur, Malaysia

[c] Guangzhou College of Technology and Business, Guangzhou, China

[d] School of Education, University of Southern California, Los Angeles, United States

Z.L.: zongrong@usc.edu; J.X.: xujunhao1105@gmail.com; S.W.: siqinwan@usc.edu; Y.W.: ywu67438@usc.edu; H.L.: lihaiyan@usc.edu

**Corresponding author:**

Siqin Wang, Spatial Sciences Institute, University of Southern California, Los Angeles, US. siqinwan@usc.edu

**Authors' contributions**

Zongrong Li: Conceptualization, Data curation, Formal analysis, Investigation, Methodology, Software, Validation, Writing – original draft

Junhao Xu: Conceptualization, Formal analysis, Methodology, Software, Validation, Writing – original draft

Siqin Wang: Conceptualization, Validation, Investigation, Resources, Writing - Original Draft, Writing - Review & Editing, Supervision, Project administration, Funding acquisition

Yifan Wu: Investigation, Writing – original draft

Haiyang Li: Software, Visualization

**Statements and Declarations**

Competing Interests

The authors declare that they have no competing financial or non-financial interests directly or indirectly related to the work submitted for publication.

**Acknowledgment**

We would like to thank Dr. Hongxu Ma from Google AI Labs for his effort in editing, commenting, and proofreading to improve the quality of this paper.

**Highlights**

We explored the potential of large language model for geospatial information predictions.

We used chain-of-thought and retrieval-augmented generation to boost model accuracy.

We conducted tri-environmental experiments across global cities for validation.

We outperformed benchmark models at least 49.43% in capturing urban characteristics.

We observed a dynamic relationship between prediction bias and urban features.

# StreetviewLLM: Extracting Geographic Information Using a Chain-of-Thought Multimodal Large Language Model

**Abstract**

Geospatial predictions are crucial for diverse fields such as disaster management, urban planning, and public health. Traditional machine learning methods often face limitations when handling unstructured or multi-modal data like street view imagery. To address these challenges, we propose StreetViewLLM, a novel framework that integrates a large language model with the chain-of-thought reasoning and multimodal data sources. By combining street view imagery with geographic coordinates and textual data, StreetViewLLM improves the precision and granularity of geospatial predictions. Using retrieval-augmented generation techniques, our approach enhances geographic information extraction, enabling a detailed analysis of urban environments. The model has been applied to seven global cities, including Hong Kong, Tokyo, Singapore, Los Angeles, New York, London, and Paris, demonstrating superior performance in predicting urban indicators, including population density, accessibility to healthcare, normalized difference vegetation index, building height, and impervious surface. The results show that StreetViewLLM consistently outperforms baseline models, offering improved predictive accuracy and deeper insights into the built environment. This research opens new opportunities for integrating the large language model into urban analytics, decision-making in urban planning, infrastructure management, and environmental monitoring.

# 1. Introduction

Geospatial predictions using machine learning have become essential across various domains, including disaster management (Mahdizadeh Gharakhanlou & Perez, 2023; Jain et al., 2020), public health (Mhasawade et al., 2021; Wiemken & Kelley, 2020), climate change (Rolnick et al., 2022), and urban planning (Tekouabou et al., 2022). Traditional machine learning has played a key role in geospatial predictions, but its limitations have become more distinct over time. One significant drawback of traditional ML is that they often rely on structured geospatial data, such as raster or vector formats, affecting their ability to handle unstructured or multimodal data (Pierdicca & Paolanti, 2022). Additionally, traditional models may face challenges in capturing complex spatial patterns and regional variations, leading to challenges with data sparsity and uneven distribution, which could affect the accuracy and generalizability of predictions (Nikparvar & Thill, 2021).

In contrast, large language models (LLMs) have shown great promise across various fields by processing vast amounts of data and reasoning across multiple modalities (Chang et al., 2024). By integrating textual, visual, and contextual information, LLMs can introduce novel covariates for geospatial predictions, thus enhancing traditional approaches. However, extracting geospatial knowledge from LLMs poses its challenges. Although using geographic coordinates (i.e., latitude and longitude) was a straightforward way to retrieve location-specific information, this approach often yields suboptimal results, particularly when dealing with complex spatial relationships and regional characteristics. As a result, the traditional model does not easily to harness the full potential of multi-modal data, hindering its effectiveness in applications demanding comprehensive, cross-modal insights.

To address these challenges, we develop StreetViewLLM, a framework that integrates multimodal data by combining LLMs with Chain-of-Thought reasoning, a method that breaks down complex tasks into smaller, logical steps. By fusing Street View Imagery (SVI) data with geographic coordinates and textual descriptions, StreetViewLLM enhances the accuracy of geospatial predictions, providing more granular insights into urban environments. SVI serves as a rich and detailed source of geospatial information that complements satellite data, offering street-level perspectives that capture the nuances of the built environment, including infrastructure, urban aesthetics, and socio-economic conditions. Additionally, the framework employs Retrieval-Augmented Generation (RAG) to refine the extraction of geographic data, seamlessly integrating textual and visual inputs to create a more comprehensive analysis.

The introduction of StreetViewLLM marks a significant advancement in geospatial analysis by combining multimodal reasoning with LLM techniques. By enabling the extraction of fine-grained geographic information from street-level images, this framework opens new avenues for urban planning, infrastructure management, and environmental monitoring. Moreover, StreetViewLLM bridges the gap in geospatial data retrieval by improving how LLMs interpret geographic coordinates concerning real-world environments, especially in lesser-known or remote locations. This innovation enhances prediction accuracy and advances the capabilities of AI-driven geospatial models, allowing geographic information to be used more effectively across a wide range of fields.

## 2. Related Work

### 2.1 Street view imagery as a promising dataset

The rise of Street View Imagery (SVI) has significantly advanced urban analytics, offering a detailed and dynamic geographical data source that complements traditional methods such as satellite imagery. SVI provides high-resolution visual data at the street level, enabling researchers to analyze and understand urban environments in greater detail. Its ability to capture fine-grained details of urban streetscapes has made SVI a powerful tool in urban studies, ranging from architectural analysis to public health assessments (Biljecki & Ito, 2021). SVI's utility has been enhanced by its integration with various urban data sources, such as street networks, building information, and socioeconomic data. For instance, Rundle et al. utilized Google Street View to audit neighborhood environments, assessing walkability, street conditions, and safety. This method allows for systematically observing built environments without requiring in-person site visits. It offers an efficient way to gather data on urban infrastructure and its impact on residents' health and well-being (Rundle et al., 2011).

Moreover, architectural analysis has benefited from the integration of SVI. Developed feature extraction module for architectural style classification using SVI data. This approach enables the classification of architectural styles based on building features captured through SVI, showcasing its potential for large-scale architectural studies (Zhao et al., 2018). Similarly, Ding and Hu explored building exterior colors and materials, using SVI to assess and analyze architectural elements at a street-level scale, providing valuable insights into urban aesthetics and design trends (Ding & Hu, 2013). SVI has also been employed in studies focused on green infrastructure and its impact on urban life; for example, the Green View Index (GVI) has been used to analyze how neighborhood greenery affects walking time. By leveraging SVI and deep learning techniques, they demonstrated that areas with higher green view indices tend to encourage more pedestrian activity, thus contributing to public health and urban planning strategies (Ki & Lee, 2021).

Furthermore, SVI has been used to measure the relationship between the built environment and crime, applying machine learning to SVI data revealing how specific urban design features, such as street visibility and building density, influence crime rates in urban areas (Hipp et al., 2021). The application of SVI also extends to global datasets, creating the Global Streetscapes dataset, which consists of 10 million street-level images from 688 cities worldwide. This dataset provides a comprehensive visual representation of urban environments across different regions, facilitating large-scale urban science research and enabling comparative analysis between cities (Hou et al., 2024). Additionally, SVI is proving useful in studies of building footprints and urban forms. Utilized satellite imagery and SVI data in tandem to extract large-scale building footprints, providing a more accurate understanding of urban development patterns (Gupta & Deb, 2023) and also explored how neighborhood walkability factors, such as street-level design and infrastructure, affect walking behaviors. Their study used a big data approach with SVI images to offer insights into how urban design can encourage or discourage physical activity. Overall, the broad application of SVI across various fields, including urban planning, architecture, public health, and crime analysis, underscores its value as a tool for understanding urban dynamics at both local and global scales. As researchers continue to harness SVI's potential for geographic detection and urban information analysis, its role in shaping urban studies will undoubtedly expand further.

*2.2 Tri-environmental Framework: Social, Built, and Natural Environments*

The Tri-environmental Framework integrates three essential dimensions of urban analysis: the social environment (population, public health), the built environment (urban infrastructure), and the natural environment (ecological systems). This framework allows researchers to evaluate urban environments holistically, selecting ground truth data from these three dimensions (Wang et al., 2023). For instance, the social environment includes metrics such as population density and public health indicators, while the natural environment uses the Normalized Difference Vegetation Index (NDVI) to measure vegetation health. The built environment often includes variables such as building height and impervious surface areas (Wang et al., 2023). The extensive population coverage and the varied information attributes of Street View Imagery (SVI) make it a practical and powerful tool for urban analysis, enabling researchers to derive more profound insights into the dynamics of urban environments, particularly as they relate to city morphology, infrastructure development, and public health. Many recent studies have increasingly utilized SVI data to gain deeper insights into urban environments. The use of SVI for geographic detection and information analysis has become a central topic of research, as it enables the detailed examination of urban features at both local and global scales. Researchers are focusing on harnessing SVI's potential for various applications, including environmental monitoring, urban planning, and public health assessments. As a result, optimizing the use of SVI for more advanced geographic detection and analytical methods continues to be a key point of interest in geospatial studies (Wang et al., 2023; Wang et al., 2023).

*2.3 Large Language Models with the Chain of Thought*

Large Language Models (LLMs) have become a powerful tool in geospatial analysis due to their ability to process and analyze vast amounts of data. However, geographic tasks often require detailed reasoning and complex problem-solving, which can challenge the typical capabilities of LLMs. To address this, researchers have introduced advanced techniques to enhance the reasoning performance of LLMs in handling such tasks. One prominent method is the Chain of Thought (CoT) approach, which improves the reasoning capabilities of LLMs by breaking down complex problems into smaller, intermediate steps. This approach has been particularly effective in tasks that require precision, enabling LLMs to perform better by logically sequencing their thought processes. For instance, the zero-shot CoT method, introduced by Wei et al., involves the addition of prompts such as "Let's think step by step," which enhances the model's ability to reason through multifaceted tasks without extensive manual inputs (Wei et al., 2022). Building upon this, researchers have begun exploring multimodal applications of LLMs, where different data types, such as textual and visual information, are integrated to improve reasoning outcomes. This Multimodal CoT framework allows LLMs to handle complex geographic tasks by utilizing various data inputs, such as satellite imagery and textual descriptions, thereby enhancing the depth and accuracy of the inferences made (Wang et al., 2024). Despite the advancements in CoT techniques, there remains a significant gap in the integration of multimodal geographic information with LLMs.

While LLMs have been used to solve geographic problems through textual and visual perspectives, these approaches have primarily been applied in isolation. For example, some studies have focused on using LLMs to interpret geographic problems from purely textual data, while others have concentrated on solving geographic tasks using visual data, such as satellite images or street view data. However, a comprehensive solution that combines both textual and visual data into a unified framework has yet to be fully developed (Wang et al., 2023; Wang et al., 2023). This paper seeks to bridge this gap by introducing a novel approach that integrates multimodal geographic information into LLMs, allowing for a more robust and sophisticated framework for geographic problem-solving. By combining textual, visual, and other forms of geographic data, this model enhances the LLM's ability to analyze complex geospatial phenomena. Although the integration of chain-of-thought reasoning and multi-modal data processing, the proposed model provides a more comprehensive understanding of geographic challenges. Comparative experiments have demonstrated that this integrated approach significantly outperforms existing models that rely on single-modal data, offering promising advancements in the fields of earth observation data and geoinformatics (Wang et al., 2024).

## 3. Methodology

In this study, we propose StreetViewLLM, a multimodal large language model framework that integrates Chain of Thought (CoT) reasoning and Retrieval-Augmented Generation (RAG) techniques to extract and utilize geographic information effectively. The research framework comprises three key operational stages, as illustrated in Figure 1: Step 1 is the geographic information retrieval, detailed in Subsection 3.1; Step 2 is the rationale generation, detailed in Subsection 3.2; and Step 3 is the answer inference, detailed in Subsection 3.3. A detailed prompt used for this process can be found in the attached Figure A 1.

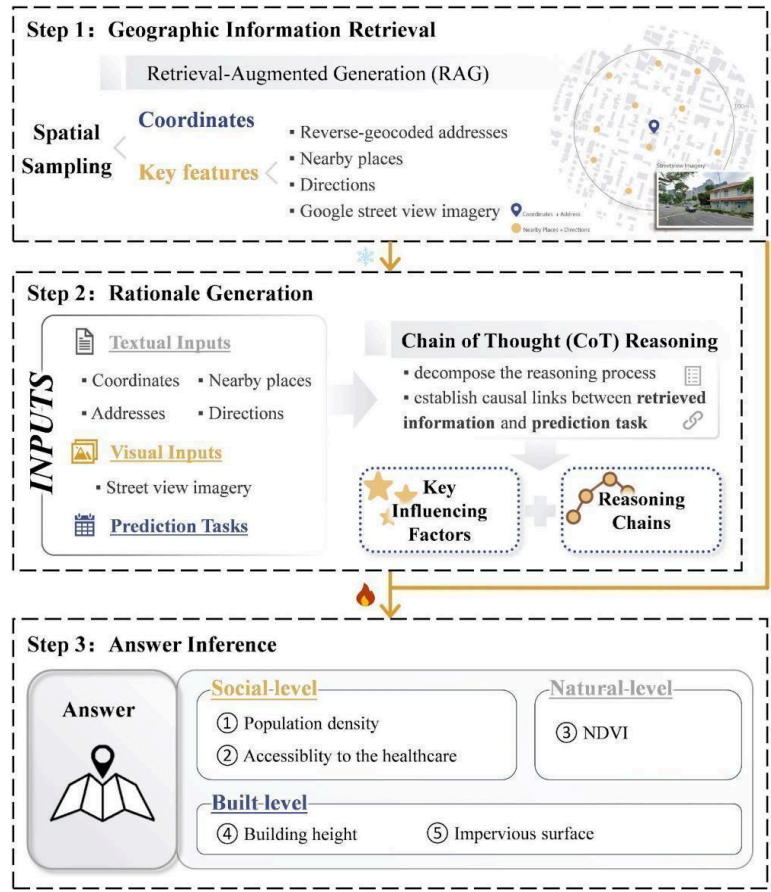

Figure 1: The Workflow of StreetviewLLM

**Note**: The snowflake icon represents zero-shot learning, while the flame icon indicates fine-tuning.

*3.1 Geographic Information Retrieval*

Retrieval-augmented generation (RAG) combines the strengths of information retrieval and generative models, which are widely used in natural language processing and dialogue systems (Lewis et al., 2021). RAG leverages an initial retrieval step to gather relevant data from extensive external databases, followed by a generative model that formulates answers based on the retrieved information. In this study, geographic RAG is applied to enhance the precision of model outputs by integrating geographically relevant data into the generation process.

We begin by conducting spatial sampling within the target city. To ensure uniformity and representativeness, we calculate the maximum distance between sampling points within the study area's bounding box to ensure uniformity and representativeness. A traversal strategy is employed, selecting sampling points from the furthest to the nearest, ensuring a well-distributed spatial sampling framework.

In addition to basic geographic coordinates, we enhance the richness of geographic information at each sampling point by collecting a set of key features, including reverse-geocoded addresses (providing detailed location descriptions), nearby places (a list of the ten closest locations within a 100-kilometer radius), and street view imagery within a 40-meter radius.

To acquire the addresses and nearby places, we employ reverse geocoding using the function/tool Nominatim (Serere, Resch, & Havas, 2023) and extract place names and locations through the Overpass API, which is an open-source API that allows querying OpenStreetMap data to obtain detailed place information efficiently (Mooney & Minghini, 2017), Street view images corresponding to the sampling points are obtained using the Google Street View API(Anguelov et al., 2010).

*3.2 Rationale Generation*

Chain of Thought (CoT) reasoning facilitates multi-step problem-solving, improving the model's ability to handle complex geographic prediction tasks (Wei et al., 2022). The core idea of CoT is to decompose the reasoning process, guiding the model through a series of intermediate steps. This is particularly beneficial in tasks where a significant gap exists between the retrieved geographic data and the prediction objective. By employing rationale generation, the model is directed to establish causal links between the retrieved information and the prediction task, enhancing both its interpretability and reasoning performance. During the rationale generation stage, we feed the model with both textual inputs (coordinates, addresses, nearby places) and visual inputs (street view imagery), along with the corresponding prediction task. This process allows the model to generate reasoning chains and identify factors that influence the prediction outcome.

*3.3 Answer Inference*

In the final stage, the rationale generated in the previous step is appended to the original input, resulting in an enriched input dataset. This augmented dataset is then fed into the answer inference model to derive the final prediction. To standardize the output, each bin is associated with a predefined value range from 0.0 to 9.9, which standardizes the output by discretizing continuous geospatial predictions into consistent, interpretable intervals (Manvi et al., 2023). ensuring consistency in the model's predictions. This range was selected based on empirical distribution characteristics, providing a balance between granularity and interpretability of the predicted results.

## 4. Experiment

*4.1 Study Area and Data*

For this project, we select seven globally representative cities—Hong Kong, Tokyo, Singapore, Los Angeles, New York, London, and Paris—due to their significant roles in the global economic, cultural, and financial landscape. Hong Kong, a major financial hub in Asia, exemplifies high urbanization and international integration. Tokyo, Japan's economic, cultural, and political nexus, is distinguished by its advanced technological infrastructure and efficient public transportation system. Singapore, serving as Southeast Asia's economic and logistical center, is recognized for its well-managed governance and highly globalized business environment. Los Angeles, the cultural and economic core of the U.S. West Coast, is renowned for its global influence in the film industry. New York, a leading global financial and cultural center, is notable for Wall Street and its multicultural demographic. London, with its rich historical significance, maintains a pivotal role as a European financial hub, while Paris, a European cultural epicenter, is distinguished by its contributions to art, fashion, and culture. These cities represent influential nodes within the global urban network, as depicted by the distribution of data points in Figure 2.

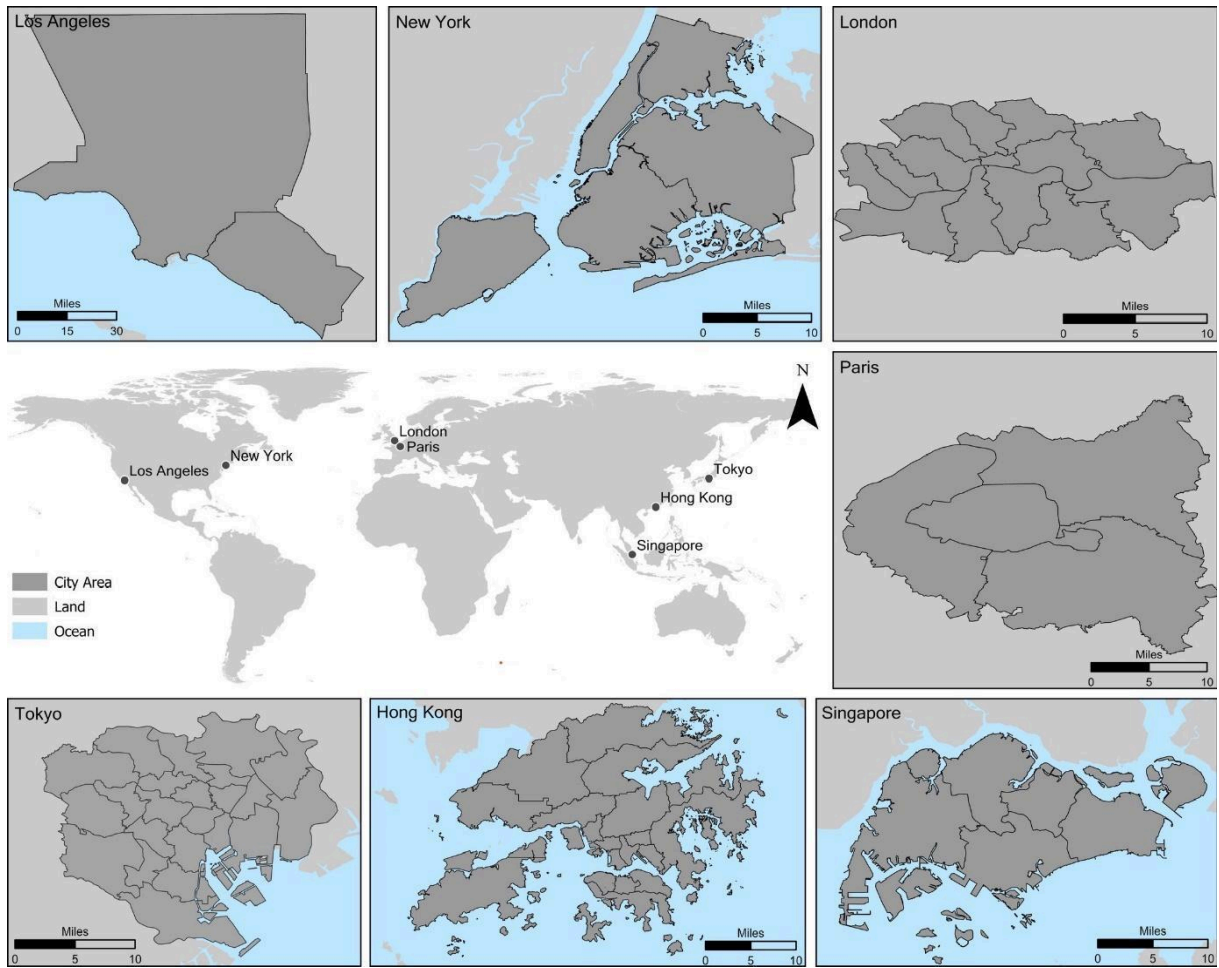

Figure 2: Study Area of StreetviewLLM

We select a wide range of geospatial prediction tasks to construct a comprehensive benchmark. The selection of tasks follows the tri-environmental framework for impact evaluation (Wang et al., 2023; Wang et al., 2023), which provides a holistic approach to understanding and assessing the impacts of environmental events by examining the interactions between the social, built, and natural environments. From these three

dimensions—social, built, and natural environments—we choose five key indicators: population density, accessibility to healthcare, Normalized Difference Vegetation Index (NDVI), building height, and impervious surface, and the specific details and measurement parameters for these indicators are provided in Table 1.

**Table 1**: Tri-Environmental Indicators for Prediction

| Dimension | Indicator | Dataset Provider | Year |
|---|---|---|---|
| **Social** | Population Density | WorldPop | 2020 |
| **Social** | Accessibility to healthcare | MAP | 2020 |
| **Natural** | NDVI | NASA LP DAAC at the USGS EROS Center | 2023 |
| **Built** | Building Height | EC JRC | 2023 |
| **Built** | Impervious Surface | EC JRC | 2023 |

**Note**: NDVI: Normalized Difference Vegetation Index; MAP: Malaria Atlas Project; EC JRC: European Commission Joint Research Centre. Details about dataset providers are provided in the reference list.

These five indicators capture the critical dimensions of the social, natural, and built environments, providing comprehensive insights into the impact of environmental events on Urban planning and green space development, like car-free cities. Population density, as a key social indicator, reflects the number of individuals residing within a given geographic area, thereby influencing resource demand and potential environmental stress. Accessibility to healthcare, another essential social metric, measures the ease with which populations can access medical facilities, a critical factor in assessing social well-being and response capabilities in the face of environmental crises. From a natural environment perspective, the NDVI serves as a measure of vegetation health, with higher values indicating more robust vegetation, which plays a vital role in mitigating urban heat and improving air quality. Within the built environment, building height offers a measure of urbanization and density, both of which are crucial for urban planning and resource distribution. Finally, impervious surface coverage identifies areas of urban development characterized by impermeable surfaces such as roads and buildings, which contribute to challenges such as urban flooding and the exacerbation of the urban heat island effect.

*4.2 Baseline*

The baseline consists of a collection of models from previous studies, ensuring that subsequent models are only considered improvements when they surpass these baselines in performance. Given the diversity and complexity of our tasks, we selected a range of the most popular machine/deep learning models. These models primarily encompass those capable of processing tabular geographic information, textual descriptions of geographic data, and street view images, as well as multimodal models that integrate geographic coordinates, text, and pictures.

1) KNN (Tamamadin et al., 2022): The K-Nearest Neighbors (KNN) algorithm is a straightforward instance-based learning method. It operates by identifying the "k" nearest data points (neighbors) to a given input using a distance metric, commonly Euclidean distance. This approach serves as a reasonable baseline, particularly for geospatial tasks, where distance metrics are highly relevant for determining spatial relationships. In our case, we have chosen k = 5.

2) XGBoost (Chen & Guestrin, 2016): XGBoost is a highly efficient and scalable gradient boosting algorithm. It serves as a robust baseline model in our analysis, offering a solid foundation for comparison. In this study, we set the number of estimators to 10.

3) MLP-BERT (Devlin et al., 2019): The MLP-BERT model is primarily designed to handle textual geographic information. It leverages BERT to encode the text, followed by a regression model based on a Multi-Layer Perceptron (MLP) to predict geographic information in our task. In this study, we utilize the 'bert-base-uncased' model for text encoding.

4) ResNet50 (He et al., 2016): Given that our dataset includes street view images, we selected ResNet50 as the foundational model for image recognition. ResNet50 is one of the most widely used and best-performing models in the field of image processing. As the original ResNet50 was designed for classification tasks, we adapted it for regression by replacing the final layer with one suitable for regression tasks. Specifically, we added a Global Average Pooling layer after the output of the pre-trained ResNet50 model. Following this setting, we introduced a fully connected dense layer with 1,024 neurons, utilizing a ReLU (Rectified Linear Unit), an activation function used in neural networks, especially in deep learning (Nair & Hinton, 2010) activation function to enhance feature learning further. ReLU is commonly used in neural networks because it helps prevent vanishing gradients and allows models to converge faster compared to other activation functions, such as sigmoid or tanh (Nair & Hinton, 2010). Finally, we added an output-dense layer with a single neuron and employed a linear activation function. Since this is a regression task, the output layer needs to return a continuous numerical prediction, making the linear activation function appropriate for this purpose.

*4.3 Sampling Data*

We collect selected geospatial data from seven globally representative cities: Hong Kong,

Tokyo, Singapore, Los Angeles, New York, London, and Paris for the experimental study. This dataset includes coordinates, addresses, nearby locations, and street view imagery, utilizing data from social (Population, Health), natural (NDVI), and built (Height, Impervious Surface) environments as ground truth indicators. In total, 10,267 data points are gathered from these cities, with a range of sample sizes. Tokyo provides the largest number of samples at 2,962, followed by Los Angeles with 2,843 and New York City with 1,448. Other cities, including Singapore and other cities, including Singapore, Paris, Hong Kong, and London, contributed fewer samples, yet they offer a diverse and robust dataset. Subsets: 60% for training, 10% for validation, and 30% for testing, providing a balanced distribution for model development and evaluation. Figure 3 illustrates the distribution of data points for the seven cities; blue represents test, yellow represents train, and red represents validation.

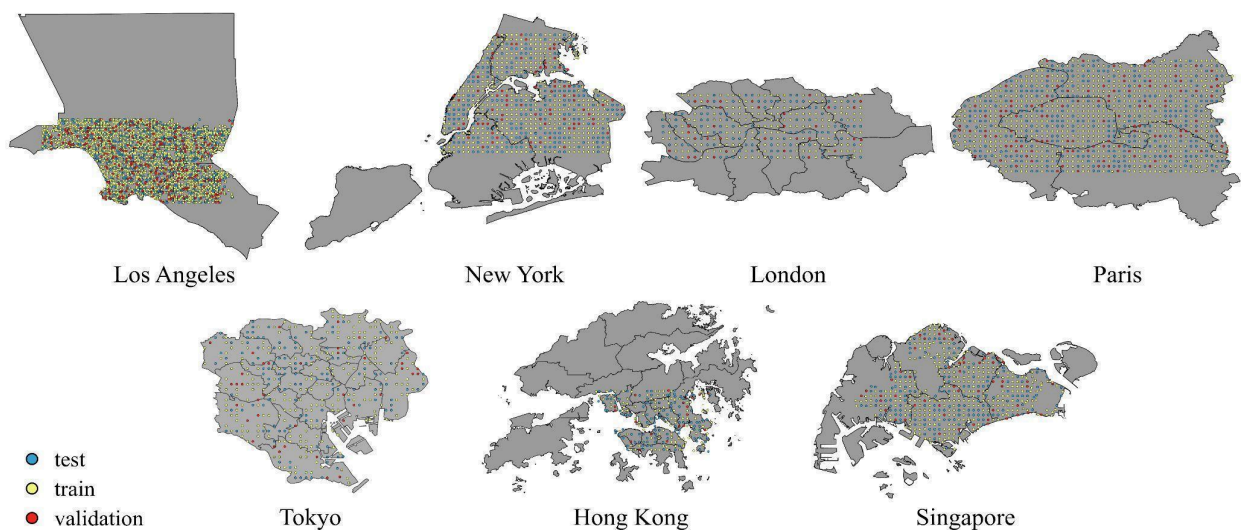

Figure 3: Spatial Distribution of Data for Model Training, Validation, and Testing in Seven Cities

**Note**: Yellow sampling points represent training data for model tuning, red sampling points represent validation data for model selection, and blue sampling points represent test data for evaluating model performance.

The sample street view from multiple cities, sourced from Google Street View, demonstrates in Figure 4 that overall sampling ensures the diversity of the data and underscores the variability and commonality in urban design and functionality across different geographical and cultural contexts. The examples from various cities illustrate their unique street characteristics. Los Angeles predominantly features single-story houses, with wide spacing between buildings and ample distance from the roads (Lehan, 2007). In contrast, New York's streets are characterized by a more significant number and density of buildings compared to other cities (Kayden, 2000). London's streets showcase a combination of multi-story buildings and broad roadways (Wang et al., 2024). In Paris, the streets are relatively narrow, a characteristic evident in both the sidewalks and the roads (Mancebo, 2020). Singapore is marked by its tall buildings and well-maintained greenery while still maintaining a significant amount of road space (Yan et al., 2022). Hong Kong, similar to Singapore, has many high-rise buildings. It also has wide roads, an extensive network of pedestrian bridges, and transportation options (Sun et al., 2019). Tokyo's streets have a more residential character, with a predominance of multi-story buildings that are denser than those in Los Angeles (Chiba et al., 2022). These street scenes reveal each city's unique response to its topographical, economic, and social pressures. This diversity is invaluable, as it allows for comparative analysis, enabling the extraction of best practices and innovative solutions applicable to multiple urban scenarios.

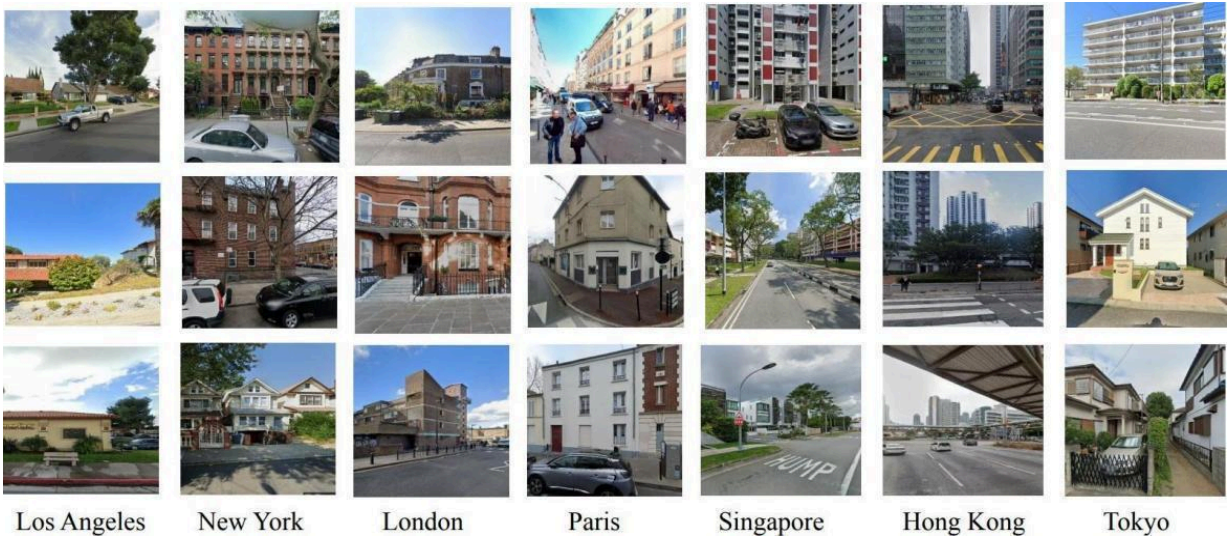

Figure 4: Samples of Street View Imagery Data

*4.4 The Performance of Model*

We perform various tasks across the social, natural, and built environments, comparing the performance of several established models against the proposed model to assess the predictive capabilities of the models. The models tested included K-Nearest Neighbors (KNN), XGBoost, MLP-BERT, and ResNet-50, in addition to our proposed model. The performance is evaluated using three key metrics: Mean Absolute Error (MAE), Root Mean Squared Error (RMSE), and R², with the latter serving as the primary measure of model fit or goodness of fit.

$$\text{Mean Absolute Error (MAE)} = \frac{1}{n}\sum_{i=1}^{n}\left|y_i - \hat{y}_i\right| \tag{1}$$

$$Root\ Mean\ Squared\ Error\ (RMSE) = \sqrt{\frac{1}{n}\sum_{i=1}^{n}\left(y_i - \hat{y}_i\right)^2} \tag{2}$$

$$R^2 = 1 - \frac{\sum_{i=1}^{n}\left(y_i - \hat{y}_i\right)^2}{\sum_{i=1}^{n}\left(y_i - \bar{y}\right)^2} \tag{3}$$

Where n is the total number of observations (data points), $y_i$ the actual value is at the i-th data point, and $\hat{y}_i$ the predicted value is at the iii-th data point.

To explore the impact of various components within the model on its predictive capability, we disassemble the model and conduct a predictive evaluation for the task of population density. The results indicate that our proposed model outperforms the comparative models across all environmental dimensions. Table 2 provides a detailed comparison of these results. This result underscores the proposed model's enhanced ability to analyze complex and less structured geospatial datasets derived from earth observation data, thereby providing more accurate predictions in natural environment variables. In the social environment predictions, which utilized indicators such as population and health, our model achieved an R² of 0.52 and 0.66, significantly higher than the next best model, KNN and XGBoost, which obtained an R² of 0.35 for population and 0.59 for Health. This is demonstrated in Table 2 by the superior capability of our model to capture the intricate relationships between social factors and urban environments. This result highlights the proposed model's enhanced ability to process and analyze natural environment variables, often characterized by more and less structured data. A complete table of predictions can be found in the attached Appendix Table A1.1-1.4.

Table 2: Performance Comparison Across Models with R²

| Task | Our Model | KNN | XGBoost | MLP-BERT | ResNet50 |
|---|---|---|---|---|---|
| Population | 0.5265 | 0.3572 | 0.3676 | 0.1752 | 0.1226 |
| Health | 0.6661 | 0.5906 | 0.4900 | 0.1613 | 0.0271 |
| NDVI | 0.5690 | 0.5693 | 0.5110 | 0.1660 | 0.0988 |
| Building Height | 0.5609 | 0.3756 | 0.3444 | 0.1745 | 0.1273 |
| Impervious Surface | 0.5224 | 0.2198 | 0.2108 | 0.0846 | 0.1348 |

## 5. Result

### 5.1. Data Summary

Following the established baseline, we propose to execute the identical task using the compiled data across a range of models, as illustrated in Table 2. Remarkably, the data from our model is significantly higher than the models that reduce the COT, Streetview, and TEXT features. In our comprehensive analysis of the built environment, incorporating critical variables such as building height and impervious surfaces, our model exhibits a superior coefficient of determination (R²) of 0.74. This performance notably surpasses that of other models, with the ResNet-50 achieving the next highest R² of 0.73, as shown in Table 3. This enhanced performance accentuates the potential of our model in proficiently tackling the task at hand, thereby indicating its viability as a preeminent solution within this domain. The model's ability to consistently outperform its counterparts underscores its effectiveness and reliability, suggesting that it holds significant promise for future applications and developments in this field.

Table 3: Modeling Performance (R²) Comparison Across Different Modules inside StreetviewLLM

| City | Our Model | WithoutCOT | WithoutStreetview | WithoutTEXT |
|---|---|---|---|---|
| Hongkong | 0.4460 | 0.4157 | 0.3420 | 0.4079 |
| Tokyo | 0.7411 | 0.7134 | 0.6755 | 0.7367 |
| Singapore | 0.5303 | 0.1983 | 0.1463 | 0.1187 |
| LA | 0.4911 | 0.2069 | 0.1096 | 0.2320 |
| NYC | 0.4500 | 0.3644 | 0.2577 | 0.2807 |
| London | 0.4341 | 0.2669 | -0.1382 | -0.1386 |
| Paris | 0.5929 | 0.5692 | 0.5748 | 0.5450 |

In our analysis, we employ the amassed data to evaluate results obtained from executing the same task across a variety of baselines. The primary criterion for this comparison is Building Height, as illustrated in the accompanying figures. Figure 5 displays the predictions for the building height metric, with green indicating better model performance, red indicating worse model performance, and gray representing areas with no data. From the examination of Figure Performance Comparison Across Different Models, it becomes evident that while the predictive performance in certain cities aligns closely with alternative models, the StreetviewLLM model exhibits a generally more efficient prediction range relative to other extant models. This finding underscores the enhanced capability of the StreetviewLLM model to achieve robust and reliable predictions across diverse urban environments.

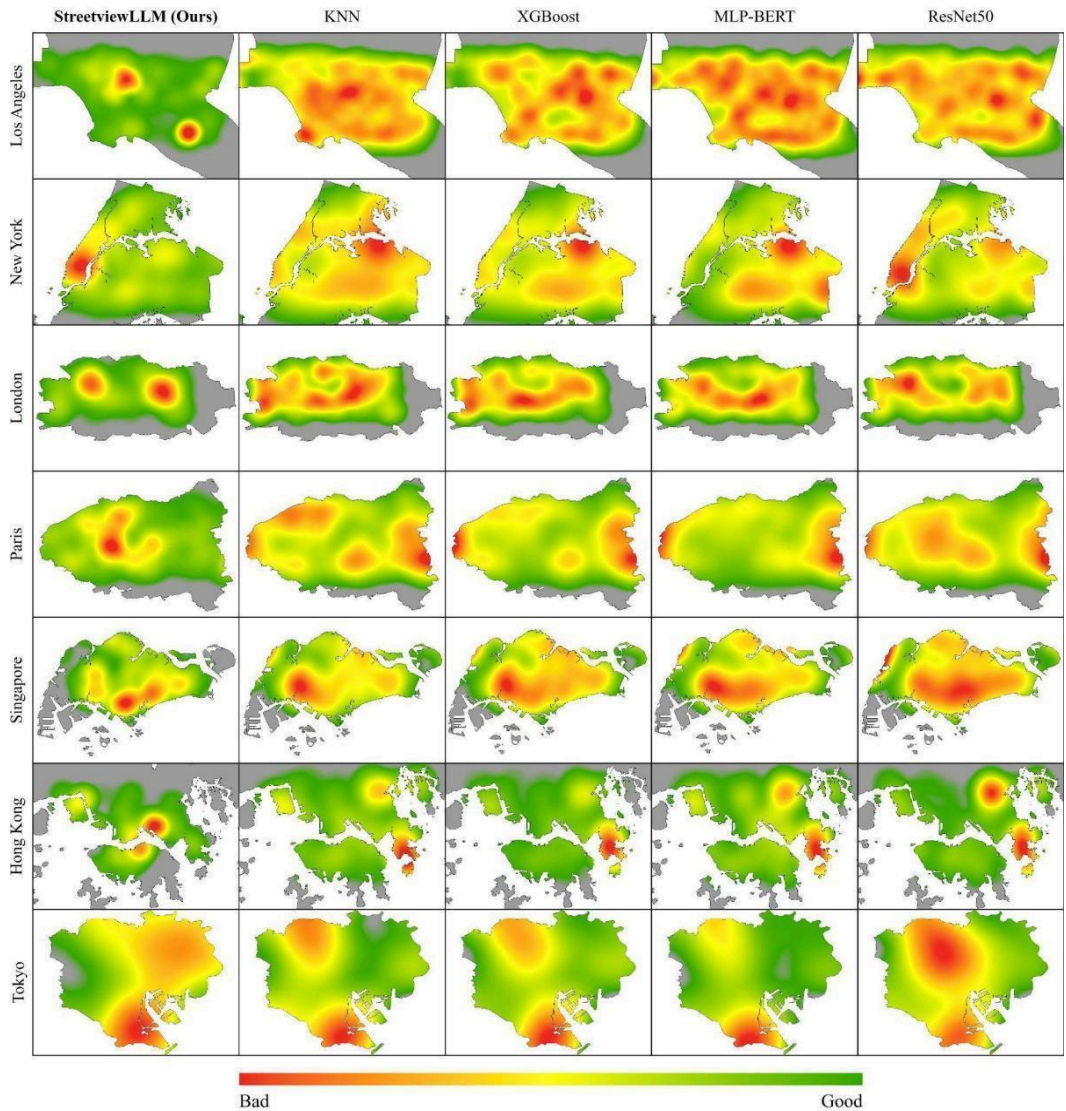

Figure 5: Performance Comparison Across Models in Building Height Task

**Note**: The color scale from red to green represents model performance, with red indicating larger differences between predicted and actual building heights (poorer performance) and green indicating smaller differences (better performance). Grey areas show regions with no available data.

This marked improvement underscores the remarkable efficacy of our model in accurately interpreting and predicting the structural characteristics of urban landscapes. Chart 3 further elucidates the comparative performance of other tested models, offering more profound insights into their respective capabilities and limitations within this analytical framework.

Overall, the proposed model consistently outperforms the other models in every dimension tested, confirming its utility in combining multimodal geographic data with a large language model (LLM) approach. The higher R² values in all three dimensions underscore its robustness and versatility in addressing complex urban and environmental challenges. These findings strongly suggest that integrating LLMs with geospatial data provides significant improvements in model accuracy and precision, far exceeding the capabilities of traditional models.

*5.2. Urban Features and Prediction Bias*

The spatial differences among cities significantly influence their urban form, ranging from high-density vertical development to expansive horizontal sprawl. Cities including Hong Kong and New York are characterized by extreme vertical density, with towering buildings and limited land space, while cities like Los Angeles are known for their sprawling, car-dependent layout. In contrast, cities such as Tokyo and Singapore balance dense development with integrated green spaces and efficient infrastructure planning. London and Paris, with their historic cores and preserved green areas, offer a mix of compact urban living and open public spaces (Rezaei & Millard-Ball, 2023; Dadashpoor et al., 2019). These diverse urban forms create unique spatial dynamics, reflecting the distinct planning strategies and geographic constraints of each city.

To explore the relationship between urban spatial features and model prediction performance, we utilized the OpenStreetMap (OSM) Overpass API to collect the number of Points of Interest (POI) within 500 meters of prediction points, categorized by different types (Residential, Commercial, Business Facilities, Industrial, Administration & Public Service, Science & Education, Green Space, and Total) (Hong & Yao, 2019). We then calculated the correlation (r²) between the POI counts and the model prediction bias. Figure 6 shows the top 10 positively correlated feature pairs (green) and the top 10 negatively correlated feature pairs (red). Further details can be found in the Appendix Table A2.

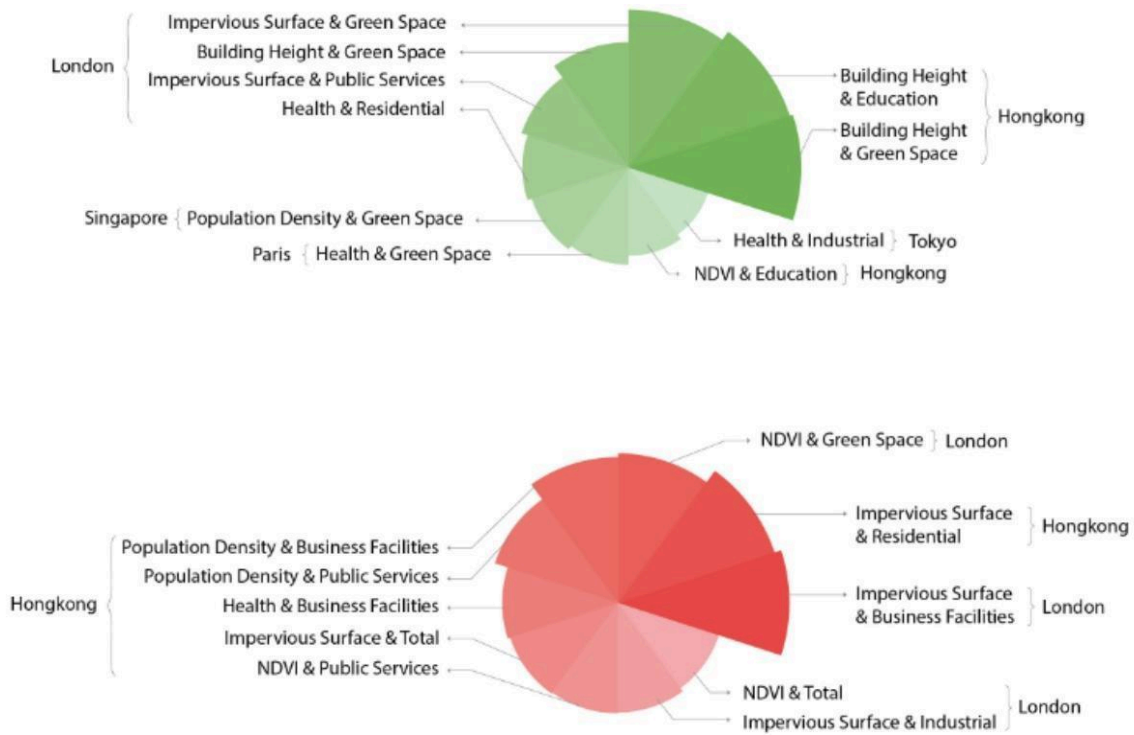

Figure 6: Correlation of Urban Spatial Features with Model Prediction Bias

**Note**: Green shows positive correlations, while red shows negative correlations across different cities. The intensity of the color represents the strength of the correlation, with darker shades indicating stronger relationships.

From Figure 6, the relationship between urban spatial features and model prediction bias varies significantly across different cities. In the top 10 positively correlated feature pairs, cities such as Hong Kong, London, Singapore, Paris, and Tokyo show notable positive correlations between certain spatial features and model bias. For instance, Hong Kong's "Building Height & Education" and "Building Height & Green Space" exhibit strong positive correlations, while London's "Impervious Surface & Green Space" is similarly positively correlated. Singapore and Paris are associated with "Population Density & Green Space" and "Health & Green Space," respectively. Tokyo's feature of "Health & Industrial" also shows a positive correlation with model bias. In contrast, the top 10 negatively correlated feature pairs show strong negative correlations in Hong Kong and London. For example, Hong Kong's "Impervious Surface & Residential" is significantly negatively correlated with model prediction bias, while London's "Impervious Surface & Business Facilities" shows a similar negative relationship. Also, Hong Kong's "NDVI & Public Services" and "Population Density & Public Services" negatively correlate with model bias.

## 6. Discussion

### *6.1 Key findings*

This study pioneers the use of multimodal large language models (LLMs) integrating with Chain-of-Thought (CoT) reasoning to predict urban features by combining visual street view imagery with textual data, including geographic coordinates, addresses, nearby places, and directions. By utilizing Retrieval-Augmented Generation (RAG) techniques, StreetViewLLM enables detailed extraction of geographic information across complex urban environments. The model demonstrates superior performance in various tasks within the tri-environment framework, covering social-level indicators (i.e., population density and accessibility to healthcare), natural-level indicators such as the normalized difference vegetation index, and built-level indicators (i.e., building height and impervious surface), outperforming traditional machine learning models like KNN, XGBoost, and ResNet50. We find that, especially in the task of predicting population density, StreetViewLLM excelled in high-density urban areas such as New York and Tokyo, where it achieved significantly higher $R^2$ values than baseline models. This outcome allows StreetViewLLM to support more precise modeling of population distributions, offering valuable insights into managing urban expansion and socio-economic planning. In terms of NDVI, the model successfully captures subtle changes in urban green spaces, with particularly strong performance in Tokyo and Hong Kong. These results open new avenues for environmental monitoring and green infrastructure development. The model also accurately predicted built-level indicators, particularly in building height and impervious surfaces. In Singapore and Paris, the model's R² surpassed other methods, positioning StreetViewLLM as a robust urban planning, infrastructure management, and environmental impact assessment tool. We further find that urban spatial layouts significantly influenced prediction accuracy: cities characterized by high vertical density, such as Hong Kong and New York, led to better predictions, while more sprawling, low-density in Los Angeles presented more significant challenges for the model.

Additionally, we observe that the model's performance was heavily dependent on integrating key modules, including Chain-of-Thought reasoning, street view imagery, and textual data. When any of these modules were removed, we saw a noticeable decrease in prediction accuracy, particularly in the $R^2$ values for various tasks. These experimental results underscore the importance of multimodal data fusion, which has proved critical for achieving the highest levels of accuracy within StreetViewLLM. Lastly, an analysis of urban features and prediction bias reveals that specific characteristics of cities, such as building density, street configuration, and green space distribution, have a measurable impact on the model's prediction accuracy. For example, cities with higher building densities, such as New York and Hong Kong, experience lower prediction bias in tasks related to building height. In contrast, cities with expansive, low-density layouts, such as Los Angeles, show higher biases in population density and healthcare accessibility predictions due to their more dispersed urban form. Furthermore, cities with well-integrated green infrastructure, such as Tokyo and Singapore, demonstrate greater accuracy in NDVI predictions, whereas areas with fragmented or limited green spaces introduce more bias. Mixed-use developments and complex street networks, particularly in cities like Paris and London, help mitigate prediction biases for population density and healthcare accessibility by providing the model with a more diverse set of data points. These findings highlight the critical role that multimodal data integration and urban features play in improving prediction accuracy, making StreetViewLLM a powerful tool for urban analytics and decision-making.

## *6.2 Contribution to the Literature*

Our experimental application of the StreetviewLLM model demonstrates its unique contribution to geospatial analysis by combining multimodal data sources for enhanced accuracy. This aligns with previous research that emphasizes the potential of Street View Imagery (SVI) in urban analysis. For instance, SVI's capacity has been highlighted to support diverse urban studies, such as infrastructure analysis and public health assessments (Biljecki & Ito, 2021). StreetviewLLM, however, advances beyond these prior studies by integrating SVI with additional geospatial data types, including population density, NDVI for natural vegetation, and built environment features like height and impervious surface measures. This multimodal integration leads to significantly higher prediction accuracy, consistent with the emphasis on multi-source data in producing robust urban environmental assessments (Gupta & Deb, 2023).

When we compare StreetviewLLM to traditional models, such as KNN, XGBoost, MLP-BERT, and ResNet50, our model consistently outperforms these benchmarks in both precision and accuracy, underscoring the advantage of LLMs equipped with diverse geospatial inputs. This improvement echoes insights the single-modal approaches often fall short in capturing complex urban interactions (Wang, et al., 2023). By leveraging multimodal data fusion, StreetviewLLM mitigates the limitations of single-modal models, which struggle to capture the depth and variability found in heterogeneous urban datasets.

Our experimental results also highlight the vital role of Chain-of-Thought (CoT) reasoning within the LLM framework. CoT enables StreetviewLLM to maintain high prediction accuracy, particularly evident in $R^2$ values, across various tasks involving intricate urban and environmental variables. Wei et al. (2022) emphasize that CoT enhances LLMs by breaking down complex questions into intermediate steps, and we observe a substantial drop in model performance when CoT is removed in baseline tests. Integrating CoT within StreetviewLLM reinforces its role in complex decision-making, aligning with Wei et al.'s findings regarding the importance of structured reasoning in improving LLM outcomes.

Our model's reliance on multimodal data integration also addresses a key gap in geospatial research, where previous LLM applications are often limited to either textual or visual data alone. Studies by Hipp et al. (2021) and Rundle et al. (2011) demonstrate the utility of individual data types—such as SVI for crime prediction and neighborhood audits—but StreetviewLLM builds on these achievements by concurrently incorporating multiple data streams. This approach allows us to produce a more holistic understanding of urban phenomena, which is particularly beneficial for applications in environmental monitoring, urban planning, and public health. StreetviewLLM could integrate with smaller models tailored to specific aspects like recent construction or traffic density in urban settings with incomplete or outdated data, such as neighborhoods undergoing rapid development. This integration would allow StreetviewLLM to refine its predictions on issues like air quality or noise pollution by drawing on precise, localized data, thus maintaining high accuracy even where comprehensive datasets are lacking.

While StreetviewLLM provides significant advances, our results also reveal areas for further refinement. For example, we find that multimodal data fusion is crucial for high accuracy but may present challenges in cases where comprehensive data across social, natural, and built environments is lacking. As a unified framework for geospatial tasks, StreetviewLLM could integrate with traditional statistical methods or smaller specialized models (e.g., single-task models for highly localized data) to handle data-scarce environments more effectively. This integration could enable the model to adapt based on available data sources, optimizing resource usage without sacrificing predictive power.

*6.3 Limitation*

Despite its novelty, this paper acknowledges several limitations, which present opportunities for future research and model refinement. The first limitation pertains to the ground truth data and the street view imagery and textual descriptions used as inputs. The current model does not account for temporal changes; indicators from different years are predicted based on the same input data, potentially introducing bias. This lack of temporal differentiation means that shifts in urban features over time are not captured, which could affect the accuracy of predictions in dynamic environments where rapid development or decay occurs (Anees et al., 2020). The use of Street View Imagery (SVI) may have introduced bias. SVI data collected by Google are typically captured by cameras mounted on vehicles, and while we apply a buffer-based resampling technique to mitigate missing street view data, some sampled regions still contain gaps. Furthermore, some SVI images, rather than capturing actual street views, include indoor scenes, which introduces noise into the model's input. Variations in the resolution, time of capture, and visual obstructions—such as weather conditions, lighting, or traffic—can also impact prediction accuracy, particularly for tasks requiring fine-grained spatial detail, like building height and impervious surface detection (Smith et al., 2021; Kang et al., 2020; Ito et al., 2024).

The second limitation lies in the integration of Chain-of-Thought (CoT) reasoning, which, while significantly enhancing the interpretability of the model, introduces additional complexity that can sometimes result in inconsistencies in the final predictions. CoT's multi-step reasoning approach is particularly beneficial for handling complex geographic tasks, as it breaks down larger problems into smaller, more manageable steps. However, this layered reasoning process can also lead to compounding errors, especially when the input data is ambiguous or incomplete. For example, minor inaccuracies in the early stages of reasoning can propagate through subsequent steps, amplifying errors in the final prediction output (Wang et al., 2022). Additionally, the CoT reasoning process comes with a high computational overhead. The model requires significant computational resources to perform the intermediate reasoning steps, increasing both the time and power needed for processing, particularly when applied to large-scale geographic datasets. This computational cost not only limits the model's efficiency but also introduces the risk of bias, as more complex reasoning chains may disproportionately affect certain types of geographic data, especially those that require intricate interpretation. As a result, while CoT improves the model's interpretability, it also poses challenges in terms of computational efficiency and prediction accuracy, especially in scenarios involving intricate geographic information. The model's generalizability across diverse urban contexts is another limitation. Although the study encompasses a variety of global cities, the training dataset primarily focuses on major urban centers, which may not fully capture the characteristics of smaller or less-developed cities. This could limit the model's applicability to environments with distinct urban forms, infrastructure, or socio-economic conditions, restricting its use in a broader range of urban settings.

Finally, while the StreetViewLLM framework demonstrates solid multimodal data integration, it faces computational challenges. The processing of large-scale datasets, particularly for real-time applications in urban planning and disaster management, requires substantial computational power and resources. This limits the model's scalability and may hinder its deployment in scenarios where quick, large-scale predictions are essential. Addressing these limitations through more temporally diverse datasets, refined data handling methods for SVI, enhanced CoT reasoning efficiency, and broader inclusion of urban environments in the training data will be crucial steps for future research.

*6.4 Future Application and Policy Implication*

This research demonstrates the significant potential of StreetViewLLM to revolutionize geospatial analysis and inform urban policy-making. By accurately predicting key urban characteristics using readily available multimodal data, StreetViewLLM can support data-driven decision-making across diverse domains.

*Smart City Design and Simulation:* Integrating StreetViewLLM with urban simulation platforms could create dynamic "digital twins" of cities. By incorporating real-time data feeds and predictive modeling, these digital twins could simulate the impact of urban planning decisions on various aspects of city life, such as traffic flow, pedestrian movement, and environmental conditions. This would allow urban planners to test different scenarios and optimize designs for sustainability, livability, and efficiency before implementing them in the real world.

*Environmental Monitoring:* The accurate prediction of NDVI empowers environmental monitoring efforts. By tracking changes in vegetation health across the urban landscape, policymakers can identify areas experiencing environmental stress and implement targeted interventions. This could include tree planting initiatives, the development of urban green spaces, and policies to mitigate air pollution.

*Disaster Response:* In disaster scenarios, StreetViewLLM can provide rapid assessments of impacted areas by analyzing street view imagery and predicting the extent of damage to buildings and infrastructure. This information can aid emergency responders in prioritizing resource allocation and optimizing evacuation strategies.

*Real Estate Appraisal and Investment:* StreetViewLLM could revolutionize the real estate sector by providing automated property valuations based on street view imagery, neighborhood characteristics, and predicted urban development trends. By analyzing factors like building height, surrounding green space, and proximity to amenities, the model could generate accurate property appraisals, aiding both buyers and investors in making informed decisions. This could also help identify undervalued properties and predict future property value appreciation, streamlining real estate investment strategies.

*Autonomous Navigation and Urban Exploration:* StreetViewLLM could enhance autonomous navigation systems by providing a deeper understanding of the urban environment. By analyzing street view imagery, the model could identify landmarks, predict pedestrian behavior, and assess the safety of different routes. This could lead to more efficient and safer navigation for autonomous vehicles and delivery robots, particularly in complex urban environments with high pedestrian density and unpredictable traffic patterns. Furthermore, the model could be used to create interactive maps for urban exploration, providing users with personalized recommendations for routes, points of interest, and hidden gems based on their preferences and the predicted ambiance of different neighborhoods.

*Policy Implications:* The insights generated by StreetViewLLM can inform the development of evidence-based urban policies. By understanding the complex interplay between urban form, environmental factors, and social indicators, policymakers can design effective strategies to promote sustainable urban development, improve public health outcomes, and enhance the resilience of cities to environmental challenges.

However, realizing the full potential of StreetViewLLM requires addressing its limitations. Future research should focus on incorporating temporal dynamics, improving the handling of noisy or incomplete street view imagery, and enhancing the efficiency of CoT reasoning. Additionally, expanding the model's training data to encompass a wider range of urban environments will enhance its generalizability and applicability to diverse urban contexts.
By addressing these challenges and continuing to refine the StreetViewLLM framework, we can unlock its transformative potential to revolutionize urban planning, promote sustainable development, and enhance the quality of life in cities worldwide.

## 7. Conclusion

In summary, our study contributes a significant advancement in the application of multimodal large language models for geospatial prediction and urban analytics. By integrating Chain-of-Thought Reasoning and Retrieval-Augmented Generation techniques, StreetViewLLM showcases the power of combining street-level imagery with geographic and textual data to offer deeper, more precise insights into urban environments. This framework not only pushes the boundaries of urban planning, infrastructure management, and environmental monitoring but also establishes a new standard for utilizing multimodal data to enhance predictive accuracy at a global scale. As cities around the world face escalating challenges such as population growth, environmental sustainability, and infrastructure demands, tools like StreetViewLLM provide a crucial pathway for informed decision-making, enabling the development of more resilient, efficient, and sustainable urban spaces. Its versatility and global applicability position it at the cutting edge of urban analytics and AI-driven geospatial research. Ultimately, StreetViewLLM holds the potential to not only influence academic research but also offer scalable and practical solutions for real-world challenges in smart cities, disaster management, and sustainable urban development, shaping the future of cities worldwide.

**Data Availability**

The analytical framework, data, and codes for data analysis in this study are publicly accessible via the project repository: https://github.com/Jasper0122/StreetviewLLM.

# Appendix A. Appendix

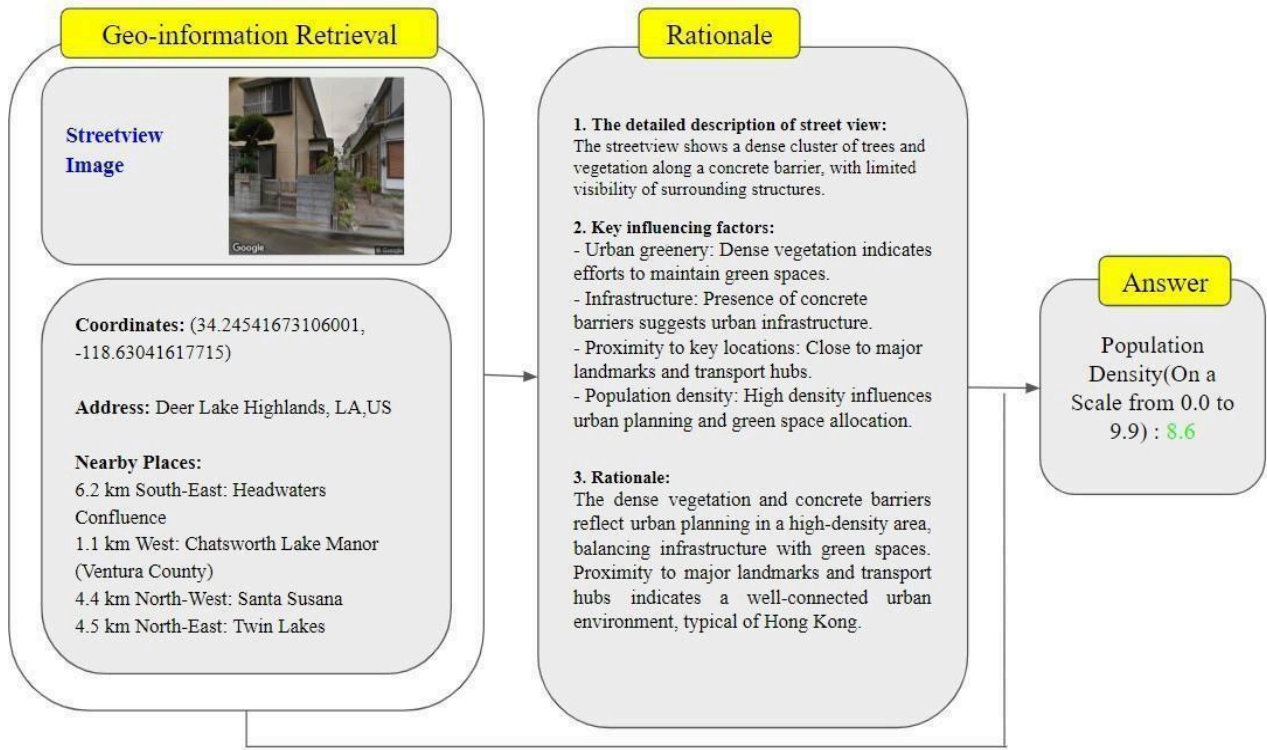

Figure A.7: Prompt Example for StreetviewLLM

Table A1.1 Performance Comparison in Population Task

| City | Our model | | | KNN | | | XGBoost | | | MLP-BERT | | | ResNet50 | | |
|---|---|---|---|---|---|---|---|---|---|---|---|---|---|---|---|
| | MAE | RMSE | R2 | MAE | RMSE | R2 | MAE | RMSE | R2 | MAE | RMSE | R2 | MAE | RMSE | R2 |
| **Hongkong** | 1.7118 | 2.3574 | 0.4460 | 1.9641 | 2.7232 | 0.2607 | 1.8609 | 2.6977 | 0.2745 | 2.0179 | 2.6153 | 0.3182 | 2.5628 | 2.9863 | 0.1110 |
| **Tokyo** | 0.6096 | 0.9772 | 0.7411 | 0.6840 | 1.0422 | 0.7050 | 0.7445 | 1.0641 | 0.6925 | 0.9318 | 1.2887 | 0.5490 | 1.2374 | 1.5479 | 0.3494 |
| **Singapore** | 1.4095 | 1.7742 | 0.5303 | 1.2282 | 2.0022 | 0.4443 | 1.2852 | 1.9822 | 0.4554 | 1.5096 | 2.0998 | 0.3889 | 1.9020 | 2.4031 | 0.1996 |
| **LA** | 1.0055 | 1.5662 | 0.4911 | 1.6071 | 2.0636 | 0.1464 | 1.6624 | 2.0166 | 0.1848 | 2.3049 | 2.8574 | -0.6353 | 2.1505 | 2.5483 | -0.3006 |
| **NYC** | 1.2755 | 2.1606 | 0.4500 | 2.2237 | 2.8013 | 0.0753 | 2.1359 | 2.6511 | 0.1718 | 1.9641 | 2.4554 | 0.2896 | 2.0994 | 2.5296 | 0.2460 |
| **London** | 1.0389 | 1.5291 | 0.4341 | 1.4116 | 1.7866 | 0.2202 | 1.2731 | 1.8045 | 0.2045 | 1.7439 | 2.1846 | -0.1659 | 1.4981 | 1.9686 | 0.0532 |
| **Paris** | 0.7332 | 1.0111 | 0.5929 | 0.6659 | 0.9414 | 0.6488 | 0.7483 | 1.0173 | 0.5899 | 0.8732 | 1.1436 | 0.4818 | 1.1100 | 1.4211 | 0.1998 |

Table A1.2 Performance Comparison in Health Task

| City | Our model | | | KNN | | | XGBoost | | | MLP-BERT | | | ResNet50 | | |
| --- | --- | --- | --- | --- | --- | --- | --- | --- | --- | --- | --- | --- | --- | --- | --- |
| | MAE | RMSE | R2 | MAE | RMSE | R2 | MAE | RMSE | R2 | MAE | RMSE | R2 | MAE | RMSE | R2 |
| Hongkong | 0.9892 | 1.5706 | 0.7513 | 1.6111 | 2.2365 | 0.4957 | 1.8538 | 2.6698 | 0.2814 | 2.1275 | 2.7954 | 0.2122 | 2.3966 | 3.2572 | -0.0695 |
| Tokyo | 0.9894 | 1.4529 | 0.6822 | 1.1270 | 1.4844 | 0.6677 | 1.2207 | 1.6020 | 0.6130 | 1.5326 | 2.0153 | 0.3876 | 1.7854 | 2.3647 | 0.1569 |
| Singapore | 1.0682 | 1.4820 | 0.6314 | 1.2327 | 1.6032 | 0.6361 | 1.3024 | 1.6021 | 0.6366 | 1.4563 | 1.9105 | 0.4832 | 1.9112 | 2.5475 | 0.0812 |
| LA | 0.9457 | 1.2258 | 0.7503 | 1.1123 | 1.4556 | 0.7095 | 1.5585 | 1.9535 | 0.4769 | 2.6408 | 3.3468 | -0.5346 | 2.3039 | 2.8243 | -0.0928 |
| NYC | 0.8745 | 1.2556 | 0.6678 | 1.0025 | 1.3165 | 0.6347 | 1.1571 | 1.5013 | 0.5250 | 1.3622 | 1.7805 | 0.3319 | 1.7006 | 2.1444 | 0.0310 |
| London | 0.6317 | 0.8603 | 0.4574 | 0.8052 | 0.9863 | 0.2730 | 0.7983 | 0.9611 | 0.3097 | 0.9545 | 1.1354 | 0.0366 | 0.9636 | 1.1533 | 0.0062 |
| Paris | 0.7527 | 1.0258 | 0.7223 | 0.8512 | 1.0346 | 0.7174 | 1.0065 | 1.2501 | 0.5875 | 1.3231 | 1.7278 | 0.2121 | 1.4249 | 0.8704 | 0.0768 |

Table A1.3 Performance Comparison in NDVI task

| City | Our model | | | KNN | | | XGBoost | | | MLP-BERT | | | ResNet50 | | |
|---|---|---|---|---|---|---|---|---|---|---|---|---|---|---|---|
| | MAE | RMSE | R2 | MAE | RMSE | R2 | MAE | RMSE | R2 | MAE | RMSE | R2 | MAE | RMSE | R2 |
| **Hongkong** | 0.8845 | 1.2364 | 0.6613 | 1.0898 | 1.4451 | 0.5673 | 1.1770 | 1.5936 | 0.4738 | 1.5515 | 1.8978 | 0.2537 | 1.5568 | 1.9879 | 0.1813 |
| **Tyokyo** | 0.5821 | 0.7967 | 0.6916 | 0.6202 | 0.8134 | 0.6783 | 0.6519 | 0.8369 | 0.6594 | 0.8002 | 1.0327 | 0.4813 | 0.9782 | 1.2321 | 0.2619 |
| **Singapore** | 0.5384 | 0.8614 | 0.5037 | 0.7203 | 0.9355 | 0.4371 | 0.7775 | 0.9850 | 0.3760 | 0.8840 | 1.1250 | 0.1861 | 0.9609 | 1.2158 | 0.0495 |
| **LA** | 0.6316 | 0.8968 | 0.5275 | 0.5770 | 0.8256 | 0.6020 | 0.6266 | 0.8585 | 0.5697 | 1.2714 | 1.6184 | -0.5271 | 1.1111 | 1.3839 | -0.1166 |
| **NYC** | 0.9888 | 1.3265 | 0.5021 | 0.7130 | 1.0199 | 0.6601 | 0.7710 | 1.1105 | 0.5971 | 0.8901 | 1.2164 | 0.5166 | 1.2308 | 1.6299 | 0.1323 |
| **London** | 0.4212 | 0.6675 | 0.5275 | 0.5360 | 0.7065 | 0.4712 | 0.5856 | 0.7590 | 0.3897 | 0.7557 | 0.9293 | 0.0853 | 0.6791 | 0.9297 | 0.0846 |
| **Paris** | 0.5010 | 0.7095 | 0.6682 | 0.5064 | 0.6456 | 0.7266 | 0.5915 | 0.7630 | 0.6183 | 0.6525 | 0.8392 | 0.5382 | 0.9141 | 1.1270 | 0.1674 |

Table A1.4 Performance Comparison in Height Task

| City | Our model | | | KNN | | | XGBoost | | | MLP-BERT | | | ResNet50 | | |
|---|---|---|---|---|---|---|---|---|---|---|---|---|---|---|---|
| | MAE | RMSE | R2 | MAE | RMSE | R2 | MAE | RMSE | R2 | MAE | RMSE | R2 | MAE | RMSE | R2 |
| **Hongkong** | 0.8446 | 1.6576 | 0.5242 | 1.5398 | 2.1604 | 0.1916 | 1.4757 | 2.0544 | 0.2690 | 1.2743 | 1.9090 | 0.3688 | 1.9478 | 2.3113 | 0.0749 |
| **Tokyo** | 0.5754 | 0.7749 | 0.6401 | 0.7374 | 0.9973 | 0.5075 | 0.7276 | 1.0145 | 0.4903 | 0.8055 | 1.1086 | 0.3914 | 0.9308 | 1.2026 | 0.2840 |
| **Singapore** | 0.8739 | 1.1929 | 0.5743 | 0.9358 | 1.3319 | 0.4671 | 1.0155 | 1.4319 | 0.3841 | 1.0275 | 1.4573 | 0.3620 | 1.2370 | 1.5768 | 0.2532 |
| **LA** | 0.4088 | 0.7377 | 0.5458 | 0.4892 | 0.7296 | 0.3660 | 0.5089 | 0.7709 | 0.2922 | 0.8634 | 1.1881 | -0.6790 | 0.7000 | 0.9712 | -0.1218 |
| **NYC** | 0.5570 | 0.7684 | 0.6398 | 0.7166 | 1.0158 | 0.3730 | 0.7341 | 0.7341 | 0.3324 | 0.7341 | 1.0293 | 0.3563 | 0.8804 | 1.1636 | 0.1775 |
| **London** | 0.4642 | 0.8273 | 0.4379 | 0.6405 | 1.0209 | 0.1438 | 0.6696 | 1.0413 | 0.1093 | 0.6522 | 1.0723 | 0.0555 | 0.6764 | 1.0788 | 0.0442 |
| **Paris** | 0.5254 | 0.8086 | 0.5640 | 0.6033 | 0.7924 | 0.5805 | 0.6156 | 0.8356 | 0.5335 | 0.7085 | 0.9737 | 0.3665 | 0.8539 | 1.1084 | 0.1794 |

Table A1.5 Performance Comparison in Impervious Task

| City | Our model | | | KNN | | | XGBoost | | | MLP-BERT | | | ResNet50 | | |
|---|---|---|---|---|---|---|---|---|---|---|---|---|---|---|---|
| | MAE | RMSE | R2 | MAE | RMSE | R2 | MAE | RMSE | R2 | MAE | RMSE | R2 | MAE | RMSE | R2 |
| Hongkong | 1.2723 | 2.6943 | 0.5410 | 3.0442 | 3.6460 | 0.1594 | 3.0463 | 3.7389 | 0.1160 | 2.8487 | 3.5565 | 0.2001 | 3.4647 | 3.6964 | 0.1360 |
| Tokyo | 1.0583 | 2.0558 | 0.4909 | 1.6306 | 2.5031 | 0.2451 | 1.6287 | 2.4774 | 0.2606 | 1.6886 | 2.6875 | 0.1298 | 1.5614 | 2.3198 | 0.3517 |
| Singapore | 1.5202 | 2.6052 | 0.5029 | 2.3104 | 3.0077 | 0.3370 | 2.3964 | 3.0820 | 0.3038 | 2.5469 | 3.1866 | 0.2558 | 2.8271 | 3.3047 | 0.1997 |
| LA | 0.7579 | 1.7066 | 0.5106 | 1.2324 | 2.0438 | 0.4304 | 1.2693 | 2.0760 | 0.4124 | 1.9759 | 3.4365 | -0.6085 | 2.0101 | 3.0269 | -0.2480 |
| NYC | 0.7636 | 1.6864 | 0.5438 | 1.5226 | 2.4369 | 0.0451 | 1.4417 | 2.3876 | 0.0833 | 1.2730 | 2.1230 | 0.2752 | 1.2743 | 2.1654 | 0.2461 |
| London | 0.6565 | 1.3394 | 0.6393 | 1.2195 | 2.1514 | 0.0693 | 1.1502 | 2.0813 | 0.1290 | 1.2982 | 2.2281 | 0.0018 | 1.1986 | 2.1822 | 0.0425 |
| Paris | 0.8762 | 2.0944 | 0.4281 | 1.4967 | 2.4120 | 0.2525 | 1.5829 | 2.5409 | 0.1705 | 1.4259 | 2.2696 | 0.3381 | 1.5928 | 2.4715 | 0.2152 |

Table A2 Correlation of Metrics with POI Columns (Positive and Negative Sorted)

| City | Metric | POI Column | Correlation with Difference |
| --- | --- | --- | --- |
| **Hongkong** | Building Height | Science and Education | 0.1934 |
| **Hongkong** | Building Height | Green Space | 0.1915 |
| **London** | Building Height | Green Space | 0.1799 |
| **London** | Impervious Surface | Green Space | 0.1439 |
| **London** | Impervious Surface | Administration and Public Services | 0.1265 |
| **London** | Health | Residential | 0.1206 |
| **Singapore** | Population Density | Green Space | 0.1154 |
| **Paris** | Health | Green Space | 0.1103 |
| **Hongkong** | NDVI | Science and Education | 0.1006 |
| **Tokyo** | Health | Industrial | 0.0923 |
| **London** | Impervious Surface | Commercial and Business Facilities | -0.2113 |
| **Hongkong** | Impervious Surface | Residential | -0.2023 |
| **London** | NDVI | Green Space | -0.1852 |
| **Hongkong** | Population Density | Commercial and Business Facilities | -0.1819 |
| **Hongkong** | Population Density | Administration and Public Services | -0.1604 |
| **Hongkong** | Health | Commercial and Business Facilities | -0.1429 |
| **Hongkong** | Impervious Surface | Total | -0.1391 |
| **Hongkong** | NDVI | Administration and Public Services | -0.1365 |
| **London** | Impervious Surface | Industrial | -0.1363 |
| **London** | NDVI | Total | -0.1311 |

# References

Anees, M. M., Mann, D., Sharma, M., Banzhaf, E., & Joshi, P. K. (2020). Assessment of Urban Dynamics to Understand Spatiotemporal Differentiation at Various Scales Using Remote Sensing and Geospatial Tools. *Remote Sensing*, *12*(8), 1306. https://doi.org/10.3390/rs12081306

Anguelov, D., Dulong, C., Filip, D., Frueh, C., Lafon, S., Lyon, R., Ogale, A., Vincent, L., & Weaver, J. (2010). Google Street View: Capturing the World at Street Level. *Computer*, *43*(6), 32–38. https://doi.org/10.1109/mc.2010.170

Biljecki, F., & Ito, K. (2021). Street view imagery in urban analytics and GIS: A review. *Landscape and Urban Planning*, *215*, 104217. https://doi.org/10.1016/j.landurbplan.2021.104217

Chang, Y., Wang, X., Wang, J., Yuan, W., Yang, L., Zhu, K., Chen, H., Yi, X., Wang, C., Wang, Y., Ye, W., Zhang, Y., Chang, Y., Yu, P. S., Yang, Q., & Xie, X. (2024). A Survey on Evaluation of Large Language Models. *ACM Transactions on Intelligent Systems and Technology*, *15*(3). https://doi.org/10.1145/3641289

Chen, T., & Guestrin, C. (2016). XGBoost: a Scalable Tree Boosting System. *Proceedings of the 22nd ACM SIGKDD International Conference on Knowledge Discovery and Data Mining - KDD '16*, 785–794. https://doi.org/10.1145/2939672.2939785

Chiba, E., Ishida, Y., Wang, Z., & Akashi Mochida. (2022). Proposal of LCZ categories and standards considering super high‑rise buildings suited for Asian cities based on the analysis of urban morphological properties of Tokyo. *Japan Architectural Review*, *5*(3), 247–268. https://doi.org/10.1002/2475-8876.12269

Dadashpoor, H., Azizi, P., & Moghadasi, M. (2019). Land use change, urbanization, and change in landscape pattern in a metropolitan area. *Science of the Total Environment*, *655*, 707–719. https://doi.org/10.1016/j.scitotenv.2018.11.267

Devlin, J., Chang, M.-W., Lee, K., & Toutanova, K. (2019). BERT: Pre-training of Deep Bidirectional Transformers for Language Understanding. *Proceedings of the 2019 Conference of the North*, *1*. https://doi.org/10.18653/v1/n19-1423

Ding, X. H., & Hu, H. H. (2013). Analysis of the Building's Exterior Color and Material. *Applied Mechanics and Materials*, *409–410*, 388–391. https://doi.org/10.4028/www.scientific.net/amm.409-410.388

Gupta, V., & Deb, C. (2023). Envelope design for low-energy buildings in the tropics: A review. *Renewable and Sustainable Energy Reviews*, *186*, 113650. https://doi.org/10.1016/j.rser.2023.113650

Haase, R., Tischer, C., & Scherf, N. (2024). Benchmarking Large Language Models for Bio-Image Analysis Code Generation. *BioRxiv (Cold Spring Harbor Laboratory)*. https://doi.org/10.1101/2024.04.19.590278

He, K., Zhang, X., Ren, S., & Sun, J. (2016). Deep Residual Learning for Image Recognition. *2016 IEEE Conference on Computer Vision and Pattern Recognition (CVPR)*, 770–778. https://doi.org/10.1109/cvpr.2016.90

Helen Ngonidzashe Serere, Resch, B., & Clemens Rudolf Havas. (2023). Enhanced geocoding precision for location inference of tweet text using spaCy, Nominatim and Google Maps. A comparative analysis of the influence of data selection. *PLOS ONE*, *18*(3), e0282942–e0282942. https://doi.org/10.1371/journal.pone.0282942

Hipp, J. R., Lee, S., Ki, D., & Kim, J. H. (2021). Measuring the Built Environment with Google Street View and Machine Learning: Consequences for Crime on Street Segments. *Journal of Quantitative Criminology*. https://doi.org/10.1007/s10940-021-09506-9

Hong, Y., & Yao, Y. (2019). *Hierarchical community detection and functional area identification with OSM roads and complex graph theory*. *33*(8), 1569–1587.

https://doi.org/10.1080/13658816.2019.1584806

Hou, Y., Quintana, M., Khomiakov, M., Yap, W., Ouyang, J., Ito, K., Wang, Z., Zhao, T., & Biljecki, F. (2024). Global Streetscapes — A comprehensive dataset of 10 million street-level images across 688 cities for urban science and analytics. *ISPRS Journal of Photogrammetry and Remote Sensing*, *215*, 216–238. https://doi.org/10.1016/j.isprsjprs.2024.06.023

Ito, K., Bansal, P., & Biljecki, F. (2024). Examining the causal impacts of the built environment on cycling activities using time-series street view imagery. *Transportation Research Part A: Policy and Practice*, *190*, 104286. https://doi.org/10.1016/j.tra.2024.104286

Jain, P., Coogan, S. C. P., Subramanian, S. G., Crowley, M., Taylor, S., & Flannigan, M. D. (2020). A review of machine learning applications in wildfire science and management. *Environmental Reviews*, *28*(4), 478–505. https://doi.org/10.1139/er-2020-0019

Jason Zhanshun Wei, Wang, X., Schuurmans, D., Bosma, M., Chi, E. H., Le, Q. V., & Zhou, D. (2022). Chain-of-Thought Prompting Elicits Reasoning in Large Language Models. *North American Chapter of the Association for Computational Linguistics*, *35:24824–24837, 2022*. https://doi.org/10.48550/arxiv.2201.11903

Kang, Y., Zhang, F., Gao, S., Lin, H., & Liu, Y. (2020). A review of urban physical environment sensing using street view imagery in public health studies. *Annals of GIS*, *26*(3), 261–275. https://doi.org/10.1080/19475683.2020.1791954

Kayden, J. S. (2000). *Privately owned public space: the New York City experience*. New York Wiley.

Ki, D., & Lee, S. (2021). Analyzing the effects of Green View Index of neighborhood streets on walking time using Google Street View and deep learning. *Landscape and Urban

*Planning*, *205*, 103920. https://doi.org/10.1016/j.landurbplan.2020.103920

Koumetio Tekouabou, S. C., Diop, E. B., Azmi, R., Jaligot, R., & Chenal, J. (2022). Reviewing the application of machine learning methods to model urban form indicators in planning decision support systems: Potential, issues and challenges. *Journal of King Saud University - Computer and Information Sciences*, *34*(8, Part B), 5943–5967. https://doi.org/10.1016/j.jksuci.2021.08.007

Lehan, R. D. (2007). *The city in literature : an intellectual and cultural history*. Univ. Of California Press.

Lewis, P., Perez, E., Piktus, A., Petroni, F., Karpukhin, V., Goyal, N., Küttler, H., Lewis, M., Yih, W., Rocktäschel, T., Riedel, S., & Kiela, D. (2021, April 12). *Retrieval-Augmented Generation for Knowledge-Intensive NLP Tasks*. ArXiv.org. https://doi.org/10.48550/arXiv.2005.11401

Mahdizadeh Gharakhanlou, N., & Perez, L. (2023). Flood susceptible prediction through the use of geospatial variables and machine learning methods. *Journal of Hydrology*, 129121. https://doi.org/10.1016/j.jhydrol.2023.129121

Mamad Tamamadin, Lee, C., Kee, S., & Yee, J.-J. (2022). Regional Typhoon Track Prediction Using Ensemble k-Nearest Neighbor Machine Learning in the GIS Environment. *Remote Sensing*, *14*(21), 5292–5292. https://doi.org/10.3390/rs14215292

Mancebo, F. (2019). Smart city strategies: time to involve people. Comparing Amsterdam, Barcelona and Paris. *Journal of Urbanism: International Research on Placemaking and Urban Sustainability*, *13*(2), 133–152. https://doi.org/10.1080/17549175.2019.1649711

Mhasawade, V., Zhao, Y., & Chunara, R. (2021). Machine learning and algorithmic fairness in public and population health. *Nature Machine Intelligence*.

https://doi.org/10.1038/s42256-021-00373-4

Milojevic-Dupont, N., & Creutzig, F. (2021). Machine learning for geographically differentiated climate change mitigation in urban areas. *Sustainable Cities and Society*, *64*, 102526. https://doi.org/10.1016/j.scs.2020.102526

Mooney, P., & Minghini, M. (2017). A Review of OpenStreetMap Data. *Mapping and the Citizen Sensor*, 37–59. https://doi.org/10.5334/bbf.c

Nikparvar, B., & Thill, J.-C. (2021). Machine Learning of Spatial Data. *ISPRS International Journal of Geo-Information*, *10*(9), 600. https://doi.org/10.3390/ijgi10090600

Pierdicca, R., & Paolanti, M. (2022). GeoAI: a review of artificial intelligence approaches for the interpretation of complex geomatics data. *Geoscientific Instrumentation, Methods and Data Systems*, *11*(1), 195–218. https://doi.org/10.5194/gi-11-195-2022

R. Swaminathan Manvi, Khanna, S., Mai, G., Burke, M., Lobell, D. B., & Stefano Ermon. (2023). GeoLLM: Extracting Geospatial Knowledge from Large Language Models. *ArXiv (Cornell University)*. https://doi.org/10.48550/arxiv.2310.06213

Rezaei, N., & Millard-Ball, A. (2023). Urban form and its impacts on air pollution and access to green space: A global analysis of 462 cities. *PLOS ONE*, *18*(1), e0278265. https://doi.org/10.1371/journal.pone.0278265

Rolnick, D., Donti, Priya L, Kaack, L. H., Kochanski, K., Lacoste, A., Sankaran, K., Ross, A. S., Milojevic-Dupont, N., Jaques, N., Waldman-Brown, A., Luccioni, A., Maharaj, T., Sherwin, E. D., Karthik, M. S., Kording, Konrad P, Gomes, C., Ng, A. Y., Hassabis, D., Platt, J. C., & Creutzig, F. (2019). *Tackling Climate Change with Machine Learning*. https://doi.org/10.48550/arxiv.1906.05433

Rundle, A. G., Bader, M. D. M., Richards, C. A., Neckerman, K. M., & Teitler, J. O. (2011). Using Google Street View to Audit Neighborhood Environments. *American Journal of Preventive Medicine*, *40*(1), 94–100. https://doi.org/10.1016/j.amepre.2010.09.034

Smith, C. M., Kaufman, J. D., & Mooney, S. J. (2021). Google street view image availability in the Bronx and San Diego, 2007–2020: Understanding potential biases in virtual audits of urban built environments. *Health & Place*, *72*, 102701. https://doi.org/10.1016/j.healthplace.2021.102701

Sun, G., Webster, C., & Zhang, X. (2019). Connecting the city: A three-dimensional pedestrian network of Hong Kong. *Environment and Planning B: Urban Analytics and City Science*, *48*(1), 239980831984720. https://doi.org/10.1177/2399808319847204

Varun Sasidharan Nair, & Hinton, G. E. (2010). Rectified Linear Units Improve Restricted Boltzmann Machines. *International Conference on Machine Learning*, 807–814.

Wang, B., Min, S., Deng, X., Shen, J., Wu, Y., Zettlemoyer, L., & Sun, H. (2022). Towards Understanding Chain-of-Thought Prompting: An Empirical Study of What Matters. *ArXiv (Cornell University)*. https://doi.org/10.48550/arxiv.2212.10001

Wang, M., Haworth, J., Chen, H., Liu, Y., & Shi, Z. (2024). Investigating the potential of crowdsourced street-level imagery in understanding the spatiotemporal dynamics of cities: A case study of walkability in Inner London. *Cities*, *153*, 105243–105243. https://doi.org/10.1016/j.cities.2024.105243

Wang, S., Cai, W., Tao, Y., Qian Chayn Sun, Pui, P., Witchuda Thongking, & Huang, X. (2023). Nexus of heat-vulnerable chronic diseases and heatwave mediated through tri-environmental interactions: A nationwide fine-grained study in Australia. *Journal of Environmental Management*, *325*, 116663–116663. https://doi.org/10.1016/j.jenvman.2022.116663

Wang, S., Cai, W., Tao, Y., Sun, Q. C., Wong, P. P. Y., Huang, X., & Liu, Y. (2023). Unpacking the inter- and intra-urban differences of the association between health and exposure to heat and air quality in Australia using global and local machine learning models. *Science of the Total Environment*, *871*, 162005.

https://doi.org/10.1016/j.scitotenv.2023.162005

Wang, Z., Xu, D., Muhammad, R., Lin, Y., Fan, Z., & Zhu, X. (2024). LLMGeo: Benchmarking Large Language Models on Image Geolocation In-the-wild. *ArXiv (Cornell University)*. https://doi.org/10.48550/arxiv.2405.20363

Wei, J., Wang, X., & Wu, C. (2022). Chain-of-thought prompting for large language models in complex decision-making. *ArXiv Preprint.* https://doi.org/10.48550/arXiv.2201.11903

Wiemken, T. L., & Kelley, R. R. (2019). Machine Learning in Epidemiology and Health Outcomes Research. *Annual Review of Public Health*, *41*(1).

https://doi.org/10.1146/annurev-publhealth-040119-094437

Yan, H., Ji, G., & Yan, K. (2022). Data-driven prediction and optimization of residential building performance in Singapore considering the impact of climate change. *Building and Environment*, *226*, 109735.

https://doi.org/10.1016/j.buildenv.2022.109735

Zhao, P., Miao, Q., Song, J., Qi, Y., Liu, R., & Ge, D. (2018). Architectural Style Classification Based on Feature Extraction Module. *IEEE Access*, *6*, 52598–52606.

https://doi.org/10.1109/ACCESS.2018.2869976